\documentclass[a4paper,conference]{IEEEtran}

\usepackage{cite}
\usepackage[pdftex]{graphicx}
\graphicspath{{figures/}}
\usepackage{amsmath}
\usepackage{url}
\usepackage{hyperref}

\usepackage{algorithm}
\usepackage{algpseudocode}

\begin{document}
\title{Temporal Alignment for History Representation in Reinforcement Learning}

\author{\IEEEauthorblockN{
Aleksandr Ermolov\IEEEauthorrefmark{1},
Enver Sangineto \IEEEauthorrefmark{2} \textsuperscript{\textsection} and
Nicu Sebe\IEEEauthorrefmark{1}}
\IEEEauthorblockA{\IEEEauthorrefmark{1}Department of Information Engineering and Computer Science (DISI), University of Trento, Italy}
\IEEEauthorblockA{\IEEEauthorrefmark{2}Department of Engineering (DIEF), University of Modena and Reggio Emilia, Italy}
\IEEEauthorblockA{Email: aleksandr.ermolov@unitn.it}}

\maketitle

\begingroup\renewcommand\thefootnote{\textsection}
\footnotetext{This work was done at DISI, University of Trento}
\endgroup

\begin{abstract}
Environments in Reinforcement Learning are usually only partially observable. To address this problem, a possible solution is to provide the agent with information about the past. However, providing complete observations of numerous steps can be excessive. Inspired by human memory, we propose to represent history with only important changes in the environment and, in our approach, to obtain automatically this representation using self-supervision. Our method (TempAl) aligns temporally-close frames, revealing a general, slowly varying state of the environment. This procedure is based on contrastive loss, which pulls embeddings of nearby observations to each other while pushing away other samples from the batch. It can be interpreted as a metric that captures the temporal relations of observations. We propose to combine both common instantaneous and our history representation and we evaluate TempAl on all available Atari games from the Arcade Learning Environment. TempAl surpasses the instantaneous-only baseline in 35 environments out of 49. The source code of the method and of all the experiments is available at \url{https://github.com/htdt/tempal}.
\end{abstract}

\IEEEpeerreviewmaketitle

\section{Introduction}
\label{Introduction}

Deep Reinforcement Learning (RL) algorithms have been successfully applied to a range of challenging domains, from video games \cite{Mnih-2013} to a computer chip design \cite{Mirhoseini2021}. These approaches use a Neural Network (NN) both to represent the current observation of the environment and to learn the agent's optimal policy, used to choose the next action. For instance, the state observation can be the current game frame or an image of the robot camera, and a Convolutional Neural Network (CNN) may be used to obtain a compact feature vector from it.

However, often RL environments are only partially observable and having a significant representation of the past may be crucial for the agent \cite{R2D2}. For instance, in the Atari 2600 Pong game, the state must be represented as at least two consecutive frames. In this way, the agent can determine both the position of the ball and the direction of its movement. More complex partially observable domains require a longer state history input to the agent. For instance, when navigating inside a 1-st person-view 3D labyrinth, the agent obtains little information from the current scene observation and needs several previous frames to localise itself.

A common solution to represent the observation history is based on the Recurrent Neural Networks (RNNs), where the RNN hidden-layer activation vector is input to the agent, possibly together with the current state observation \cite{A3C}. However, RL is characterized by highly nonstationary data, which makes training unstable \cite{DBLP:journals/corr/SchulmanMLJA15} and this instability is exacerbated when a recurrent network needs to be simultaneously trained to extract the agent's input representation. In this case, the network must obtain the representation of multiple high-dimensional input frames supervised with only the reward signal. This results in a long, very computationally demanding training.

In this paper, we propose a different direction, in which the agent's observation history is represented using compact embeddings, which describe a general state of the world. Environment-specific embeddings are trained in parallel with the RL agent using the common self-supervised contrastive learning method. Being low-dimensional, the embeddings are input directly to train the policy, indicating the most important changes in the past. Intuitively, this is inspired by the common human behaviour: when making decisions, humans do not keep detailed visual information of the previous steps. For instance, while navigating through the halls of a building, it is sufficient to recall a few significant landmarks seen during the walk, e.g. specific doors or furniture.

Following this idea, in this paper, we propose a Temporal Alignment for History Representation (TempAl) approach, composed of 3 iterative stages: {\em experience collection, temporal alignment} and {\em policy optimisation}. We obtain the representation by aligning past state observations using the pairwise cross-entropy loss function \cite{CPC,PCE}. In more detail, we minimize the distance between the latent representations of temporally-close frames. This is a form of {\em self-supervision}, in which no additional annotation is needed for the representation learning, the network is trained to produce consistent embeddings for semantically similar input (positives), contrasting them with other, more distant samples (negatives) \cite{CPC,ST-DIM,moco,simclr}.

Once the observations have been aligned, the embedding of a given frame is used as the semantic representation of the observation and is recorded in a longer {\em history}. Finally, the history is input to the agent together with an {\em instantaneous} observation representation, obtained, following \cite{Mnih-2013}, as a stack of the last 4 frames. This information is now used for policy optimisation. Note that our proposed history representation is independent of the specific RL approach. However, in all our experiments we use the PPO algorithm \cite{PPO} for the policy and the value function optimisation. The 3 stages are iterated during training, and thus past embeddings can be modified and adapted to address new observations, while the agent is progressing through the environment.

We summarize the contribution of this paper below. First, we use the pairwise cross-entropy loss function \cite{CPC,PCE} to align the agent's past observations and to obtain low-dimensional embeddings. Second, we represent the state of the agent with multiple previous observations as embeddings, so far trained. Finally, we combine both instantaneous and history representation and we evaluate our method using the PPO algorithm on all available 49 Atari games from the Arcade Learning Environment (ALE) \cite{ALE} benchmark. Our experiments confirm that this combination provides a significant boost with respect to the most common instantaneous-only input. Specifically, we show that TempAl outperforms the baseline in cases, where the history can increase observability. The source code of the method and of all the experiments is available at \url{https://github.com/htdt/tempal}.

\section{Related Work}
\label{Related}

\subsection{Partially Observable Markov Decision Processes}
The goal of the agent is to maximize a cumulative reward signal and it interacts with the environment to achieve it. The interaction process is defined as a Partially Observable Markov Decision Process (POMDP) $<S, A, T, R, \Omega, O, \gamma>$ \cite{Monahan,Kaelbling}, where $S$ is a finite state space, $A$ is a finite action space, $R(s, a)$ is the reward function, $T(\cdot|s, a)$ is the transition function, $\Omega$ is a set of possible observations, $O$ is a function which maps states to probability distributions over observations and $\gamma \in [0, 1)$ is a discount factor used to compute the cumulative discounted reward. Within this framework, at each step $t$, the environment is in some unobserved state $s \in S$. When the agent takes an action $a$, the environment transits to the new state $s'$ with probability $T(s' | s,a)$, providing the agent with a new observation $o' \in \Omega$ with probability $O(o'|s')$  and reward $r \sim R(s, a)$. 

POMDPs are an active research area and several approaches have been recently proposed to address partial observability. DQN \cite{Mnih-2013} is one of the earliest deep RL models directly trained using the high-dimensional observation $o_t$ as input (i.e., the raw pixels of the current frame). To deal with partial observability, the input of the model is composed of a stack of the 4 last grayscale frames, processed as 4 channels using a 2D CNN. We also use this observation representation which we call {\em instantaneous} to highlight the difference from the past observation history.

 \cite{DRQN} replace the fully-connected (FC) layer in DQN with an LSTM \cite{DBLP:journals/neco/HochreiterS97} layer. At each step, the model receives only one frame of the environment together with the LSTM hidden-layer value which represents the past. The idea was extended in \cite{DVRL} by inferring latent belief states. To simulate reduced observability and demonstrate the robustness of the method, \cite{DRQN} introduce the flickering Atari games, obscuring the observation with a certain probability at each step. These methods focus on  short-term partial observability, such as noisy observations, and approximate the state with an implicitly inferred continuous history. In contrast, our method explicitly represents each step of the past using an independently trained encoder, and concatenates a sequence of steps into a history. Our final goal is to augment the current state representation with important past changes of the environment.

Recurrent layers (e.g. LSTM or GRU \cite{GRU}) are a common choice for modern RL architectures. The A3C algorithm \cite{A3C} showed a high mean performance with an LSTM layer. R2D2 \cite{R2D2} is specifically focused on processing long-history sequences (80 steps with additional 40 burn-in steps) using an LSTM. However, despite its widespread, RNNs have disadvantages within an RL framework because they increase the training complexity (Sec.~\ref{Introduction}). In this paper, we propose a compact history representation that can be processed with much simpler networks. In our experiments, we use a small 1D CNN with kernel size $1$, which has only $68$ trainable parameters. We empirically show in Sec.~\ref{Results} that our proposal improves the performance for most of the environments.

\subsection{Self-supervised Learning and Metric Learning}
In this section we present related works focused on representation learning. In self-supervised learning (SSL), the network is trained to extract useful representations from data (e.g. images, videos) based on a {\em pretext task} without using annotation. In RL, observations are obtained dynamically, thus labeling them is impractical, while a suitable pretext task may require minimal effort. CPC \cite{CPC} is a pioneering SSL method, that was demonstrated on natural images and in the RL domain. For images, the method predicts the representation of a patch given a context obtained from other patches. For RL, the method predicts the representation of an observation given previous observations in a sequence, and the SSL loss is combined as auxiliary with the RL loss. Similarly, our method uses temporal positions in the trajectory. However, we explicitly align the latent representation instead of prediction, and we do not combine losses, performing each stage separately.

\cite{CPC} have introduced the InfoNCE loss, which became a foundation for several popular methods. E.g., MoCo \cite{moco} and SimCLR \cite{simclr} use this loss to align representations of two versions of one image, produced by random augmentations (cropping, color jitter etc.). Later, several alternative non-contrastive losses were proposed \cite{byol,W-MSE,SimSiam,barlow}. However, in our setting InfoNCE provides better stability, we employ it for our experiments. In parallel with SSL, this loss was developed for metric learning, it was introduced as N-pair loss \cite{NPair} and pairwise cross-entropy loss \cite{PCE} emphasizing the connection with the cross-entropy loss for supervised classification.

Following CPC concepts, ST-DIM \cite{ST-DIM} used both patch structure and temporal positions of observations, aligning nearby frames. The method was validated on several Atari games from ALE measuring if the specific information from the environment is encoded in the representation; the RL task was not considered. Recently, \cite{schwarzer2021dataefficient} have shown that the self-supervision from temporal positions and from augmentations can improve data efficiency in RL, using BYOL \cite{byol} as an auxiliary loss.

SSL is widely adopted in world models. The goal of such models is to capture world dynamics, and it is often unreasonable to spend the model capacity predicting all pixel-level details. Thus, in World Models \cite{world-models}, observations are initially encoded with VAE \cite{VAE} and the model operates in the resulting low-dimensional space. \cite{LWM} have utilised temporal alignment of observations to learn latent space, and the world model approximates dynamics in this space to address the exploration problem. \cite{EffZero} extend model-based MuZero \cite{muzero} with the self-supervised consistency for nearby observations based on SimSiam \cite{SimSiam} loss, demonstrating  the high sample efficiency. Differently from these works, TempAl is used in the model-free setting.

Our representation learning process can be viewed from the metric learning perspective. Metric is defined as a distance between objects representing their specific similarity. At the same time, TempAl representation indicates the temporal distance between observations. This idea was developed in Time-Contrastive Networks \cite{sermanet2018timecontrastive}. To obtain positives, one of the versions of the method used the temporal window as in \ref{tempal}. The performance was demonstrated in a robotic imitation setting. Moreover, several prior works \cite{slow_feat,goroshin2015unsupervised} used temporal coherence for representation learning and our method can be seen as a continuation of this line of research.

\section{Temporal Alignment and History Representation}

\subsection{Temporal Alignment}
\label{tempal}

\begin{figure}[h!]
\centering
\includegraphics[width=0.45\textwidth]{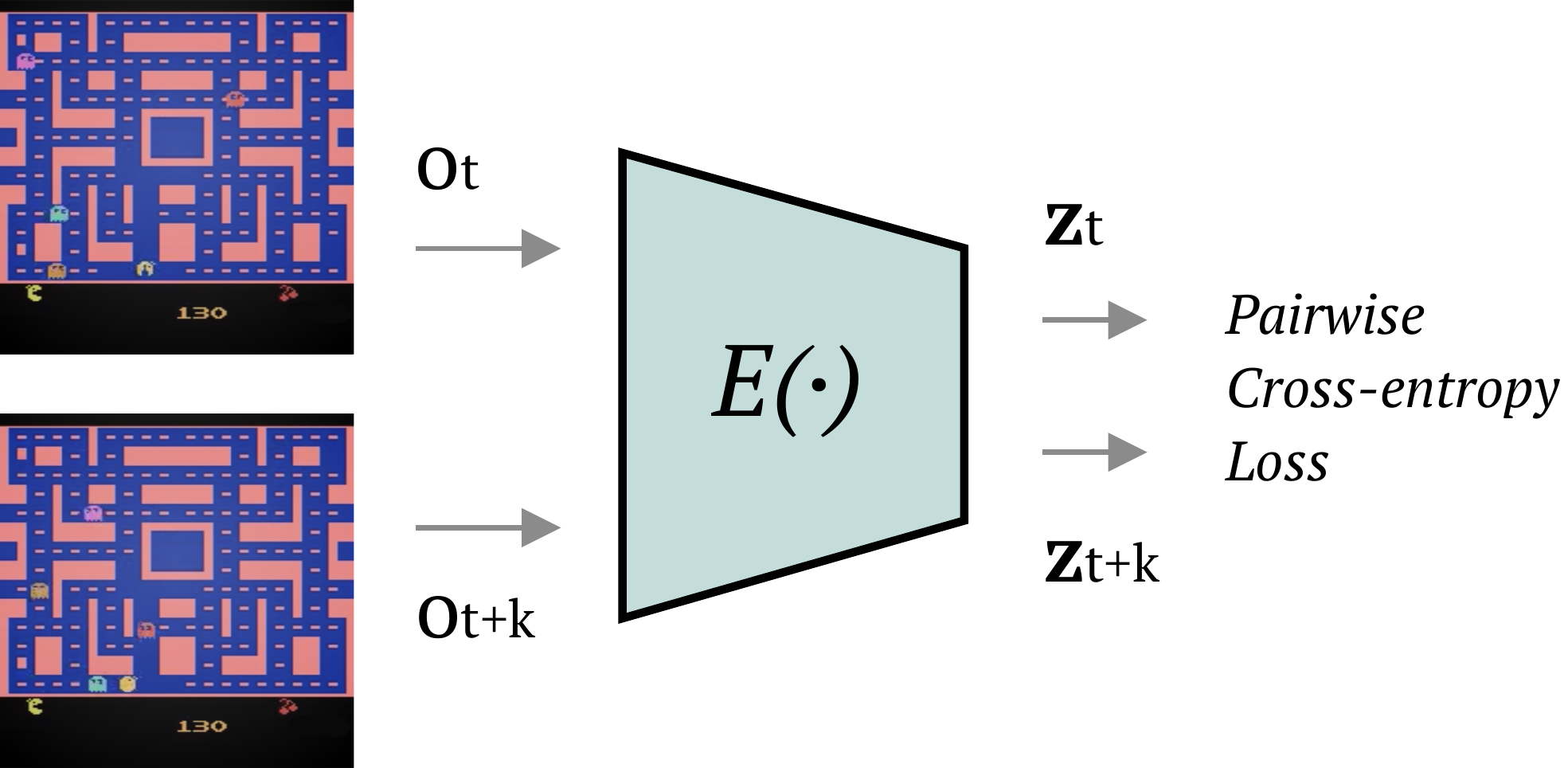}
\caption{Scheme of the temporal alignment. First, two frames are sampled within the temporal window $L$ from one trajectory. They are encoded with $E(\cdot)$, which consists of a CNN with an FC layer and $L_2$ normalization. Finally, the resulting embeddings $\mathbf{z}_t$ and $\mathbf{z}_{t+k}$ are pulled closer to each other, while other samples in the batch are pushed away using the pairwise cross-entropy loss.}
\label{fig.encoder}
\end{figure}

As mentioned in Sec.~\ref{Introduction}-\ref{Related}, for our self-supervised representation learning, we use the {\em pairwise cross-entropy loss}, which is a variant of the {\em contrastive loss} \cite{DBLP:conf/cvpr/HadsellCL06}. Following the common formulation proposed by \cite{CPC}:

\begin{equation}
\label{eq.xent}
    L_{cross-entropy} = - \log \frac{\exp{(\mathbf{z}_i^T  \mathbf{z}_j / \tau)}}{  \sum_{k=1, k \neq i}^K \exp{(\mathbf{z}_i^T  \mathbf{z}_k / \tau) } },
\end{equation}
\noindent
where $\mathbf{z}$ is a latent-space representation of corresponding observation $o$, indexes $i$ and $j$ indicate a {\em positive} pair, $K$ is the size of the current batch, $\tau$ is a temperature hyperparameter. Temporally-close observations share the same general state of the environment, while distant observations are likely to represent different states. Following this assumption, we use nearby observations as positives.
Let $o_t$ be the observation of the environment at time step $t$, presented as an image. We want to learn a representation mapping $E(o_t) = \mathbf{z}_t$, $E(\cdot)$ is our {\em self-supervised encoder} and it is based on a CNN \cite{Mnih-2013} followed by a FC layer with $L_2$ normalization (Fig.~\ref{fig.encoder}). We use two observations extracted from the same trajectory: $o_t$ and $o_{t+k}$, where $k \sim U[1,L]$. $L$ defines a temporal translation window in which observations are assumed to most likely have the same semantic content. Small values of $L$ (e.g. $L = 1$) reduce the uncertainty and thus improve the stability during training. However, if this value is too small, positives can often be identical and the encoder would be trained to simply detect duplicates. On the other hand, larger values of $L$ produce a higher variability, in which the representation learning process is forced to extract more abstract semantics to group similar frames. During preliminary experiments, we observed that the optimal value lies in the range $(1, 10)$ depending on the environment. On average, value $L=4$ provides the best result, thus we use it in our experiments in Sec.~\ref{Results}. However, it's possible to tune this parameter independently for each environment. Another parameter is the representation dimensionality $dim(\mathbf{z})$. We set it to $32$ for all experiments, in considered cases it is sufficient.
Additionally, inspired by the commonly used random crop augmentation in self-supervised learning \cite{simclr}, we randomly shift the image vertically and horizontally by the value from the range $(-4, 4)$, improving the encoder robustness.

\subsection{History Representation and Policy Optimisation}

\begin{figure*}[h!]
\centering
\includegraphics[width=0.85\textwidth]{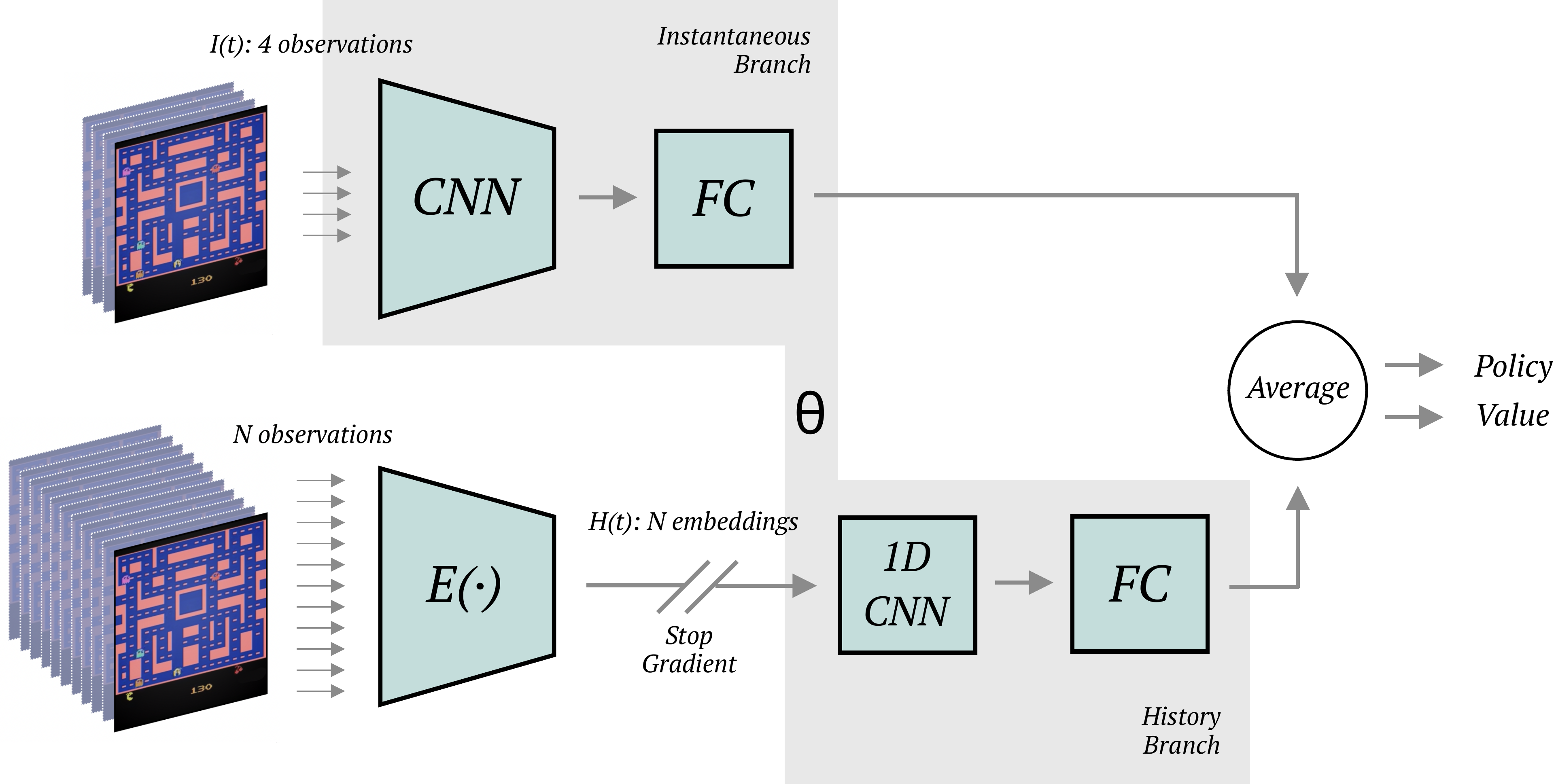}
\caption{Scheme of the RL agent architecture. The NN $\Theta$ consists of two branches: instantaneous and history. The former is a common RL architecture, which processes four recent observations and outputs policy probabilities and the state value. The latter is designed to process embeddings of $N$ history steps. First, these embeddings are obtained from the encoder $E(\cdot)$. Next, they are processed in the history branch, which predicts the policy and value in the same format as the instantaneous branch. Finally, both branches are ensembled and the output is optimized with the PPO algorithm. Note that in this scheme only the NN $\Theta$ is updated, the encoder $E(\cdot)$ is frozen during the policy optimisation stage.}
\label{fig.agent}
\end{figure*}

Once $E(\cdot)$ is updated (see Sec.~\ref{Training}), we use $\mathbf{z}_t = E(o_t)$ to represent an observation at time $t$ and update the agent's policy. In more detail, we represent the agent with an Actor-Critic architecture approximated with a NN $\Theta$. We use the PPO \cite{PPO} algorithm to optimise the policy and the value function. At time step $t$, the agent receives two types of data as input, an {\em instantaneous} representation $I(t)$ consisting of 4 concatenated raw-pixel, grayscale frames $I(t) = \{o_t, o_{t-1}, o_{t-2}, o_{t-3} \}$ and a {\em history} computed with $E(\cdot)$ over the past $N$ steps: $H(t) = \{ \mathbf{z}_t, ..., \mathbf{z}_{t-N+1} \}$. $\mathbf{z}_t$ is a low-dimensional vector ($32$), and the history size $N$ can be set to a specific value depending on the environment (we use $N=16$), thus $H(t)$ is a low-dimensional ($32 \times 16$) matrix which represents significant information about the recent agent's past trajectory. Importantly, in contrast with an RNN-based history representation, which may be prone to forgetting past information, in our proposed history representation $H(t)$, all the past $N$ observations are stored, together with their time-dependent evolution.
 
The NN $\Theta$ consists of two branches with respect to the input: instantaneous and history. The instantaneous branch consists of a NN, introduced by \cite{Mnih-2013} and commonly used for Atari in several works, including our baseline PPO. In the history branch the input $H(t)$ is first passed through a 1D CNN with the kernel size $1$, then flattened and processed with two FC layers. In this branch, we use two sub-networks with described architecture, one for the policy and one for the value. Finally, outputs of both branches are combined using an average. Such a basic ensembling allows the agent to choose a suitable input for a concrete task. We also include ablation studies (\ref{ablation}) with only the history branch. Fig.~\ref{fig.agent} shows the full architecture of our agent.

We highlight that the self-supervised encoder $E(\cdot)$ does not share parameters with the RL network $\Theta$, and it is not trained during the policy optimisation stage. Moreover, while the $E(\cdot)$ encoder is trained in a self-supervised fashion, the CNN in $\Theta$, used to extract representation from $I(t)$, is trained {\em using the RL reward}. This emphasizes the complementary between the knowledge of the two networks, being the former based on extracting visual similarities in data that do not depend on the specific RL task, while the latter being trained as in a standard RL framework, focusing on details important for action selection.

\section{Experimental Evaluation}

\subsection{Training}
\label{Training}

Initially, we run a random agent for a small number of steps ($1\%$ of the total budget) storing all obtained observations. Next, we train the $E(\cdot)$ encoder using these observations for $n_{pretrain}$ epochs. After this pretraining, the RL agent can use $E(\cdot)$ to encode the history and it is ready for the main training loop. During it, we run the RL agent in several parallel environments for a predefined number of steps (one rollout). The actions are chosen according to the current policy ({\em observation collection stage}). Next, using the collected experience, we update $E(\cdot)$ for $n_{encoder}$ epochs ({\em temporal alignment stage}) and $\Theta$ for $n_{policy}$ epochs ({\em policy optimisation stage}). These three stages are iterated, so to adapt $E(\cdot)$ to the observation distribution change.

\subsection{Implementation Details}
\label{Implementation}

\begin{table}
\renewcommand{\arraystretch}{1.5}
\caption{Configuration}
\label{table.cfg}
\centering
\begin{tabular}{|l|c|}
\hline
Embedding size ($dim(\mathbf{z})$) & 32 \\
History size $N$ & 16 \\
Temporal window $L$ & 4 \\
Maximal spatial shift & 4 \\
Number of epochs $n_{pretrain}$ & 2500 \\
Number of epochs $n_{encoder}$ & 1 \\
Number of epochs $n_{policy}$ & 3 \\
Batch size (temporal alignment) & 2048 \\
Learning rate (temporal alignment) & $2\times10^{-4}$ \\
Temperature $\tau$ & 0.1 \\
Observations storage size & 5120 \\
1D CNN channels ($\Theta$ history branch) & 4 \\
FC size ($\Theta$ history branch) & 128 \\
\hline
\end{tabular}
\end{table}

The weights of $E(\cdot)$ and $\Theta$ are initialized with a (semi-) orthogonal matrix, as described in \cite{ortho_init}, and all the biases are initialized with zeros. We use one network architecture and one set of hyperparameters for all our experiments. We adopt the training procedure and the hyperparameters in \cite{PPO}, including the total number of time steps equal to 10M. Our agent network architecture is similar to \cite{PPO}, except for the additional history branch. Encoder $E(\cdot)$ is composed of a CNN and a FC layer (Sec.~\ref{tempal}). Specifically, its CNN configuration is similar to \cite{Mnih-2013}, except the first layer, which takes only one channel as input, which corresponds to one grayscale observation. In Tab.~\ref{table.cfg} we list all other hyperparameters.

Pairwise cross-entropy loss makes use of batch samples as negatives and the sampling procedure can affect the optimization process. If observations are sampled from very distant parts of the trajectory, separating them is trivial. And vice versa: batch sampled from a small window makes the problem unsolvable. In all our experiments we sample pairs from $5$ last rollouts, each rollout contains $128$ steps of $8$ parallel environments.

\subsection{Environments}
We evaluate TempAl on the challenging deep RL benchmark ALE \cite{ALE}. Following closely the experimental protocol of the baseline \cite{PPO}, we use all 49 available classic Atari 2600 video games. In these environments, $o_t$ is an RGB image. The action space is discrete and the number of actions varies in each game from 4 to 18. We apply the same observation pre-processing (frame downscaling, etc.) as in \cite{PPO}.

\subsection{Results}
\label{Results}

\begin{figure}[ht!]
\centering
\includegraphics[width=0.49\textwidth]{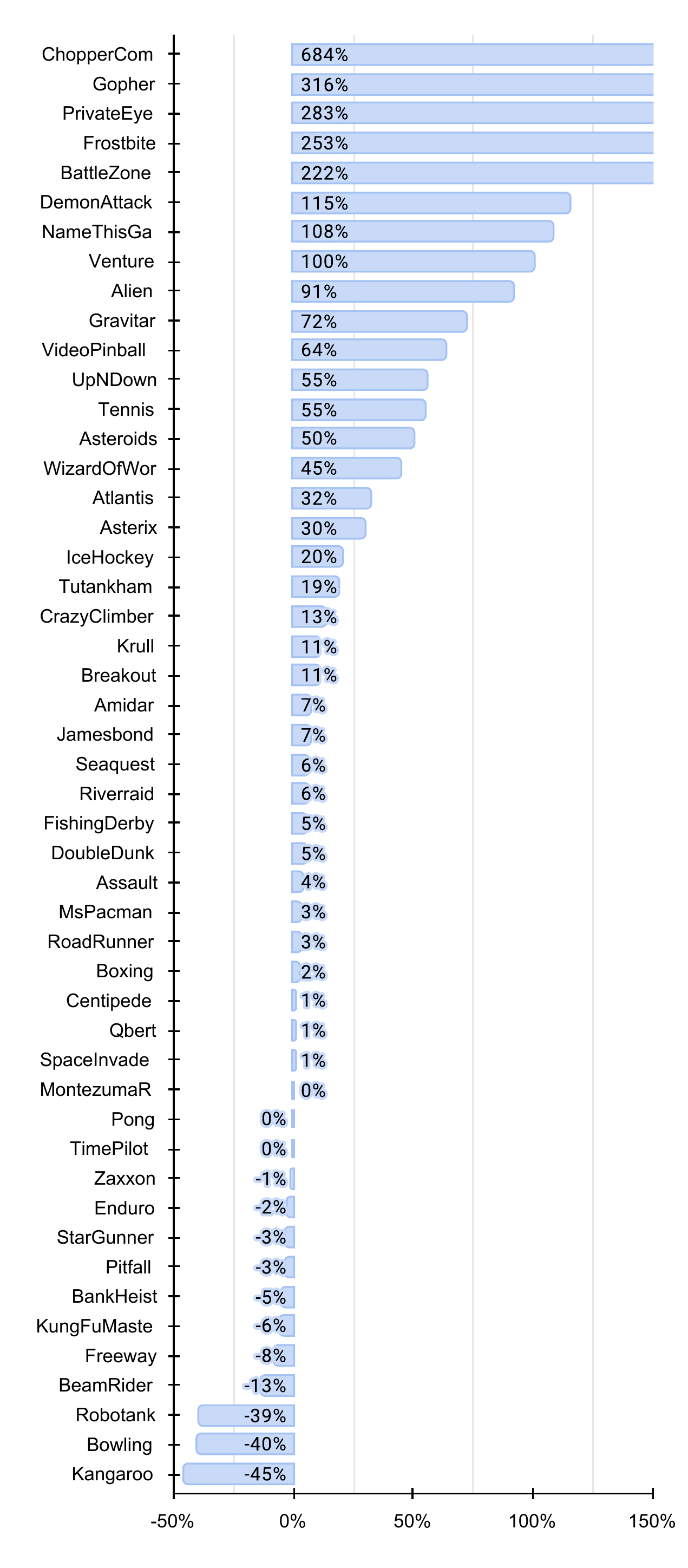}
\caption{Relative improvement (Eq.~\ref{eq.eval}) of TempAl with respect to the baseline architecture with only the instantaneous input. $0\%$ indicates that our method obtains the same cumulative reward as the baseline, while higher values indicate superior performance. Note that for \texttt{Venture} $R_{random} = R_{instant} = 0$, while $R_{TempAl} = 35$, the improvement is shown as $100\%$.}
\label{fig.results}
\end{figure}

In all the experiments the scores are averaged over 100 episodes. Fig.~\ref{fig.results} shows the relative improvement with respect to baseline \cite{PPO}. It is calculated as
\begin{equation}
\label{eq.eval}
\frac{R_{TempAl} - R_{random}}{R_{instant} - R_{random}} - 1,
\end{equation}
\noindent
where $R_{TempAl}$ is the final cumulative reward of our method, $R_{random}$ is the reward of a random agent, and $R_{instant}$ is the reward of the baseline \cite{PPO}. We can see that there is performance improvement in 35 out of 49 environments. A small difference with the baseline indicates that the game is likely reactive and the history does not add information. Large negative values indicate failure cases (3 environments), where the history input interferes with the instantaneous input. Finally, for 14 environments we can see a strong improvement ($>50\%$), in these cases, the history provides a clear advantage for the RL algorithm.

\begin{figure}[h!]
\centering
\includegraphics[width=0.47\textwidth]{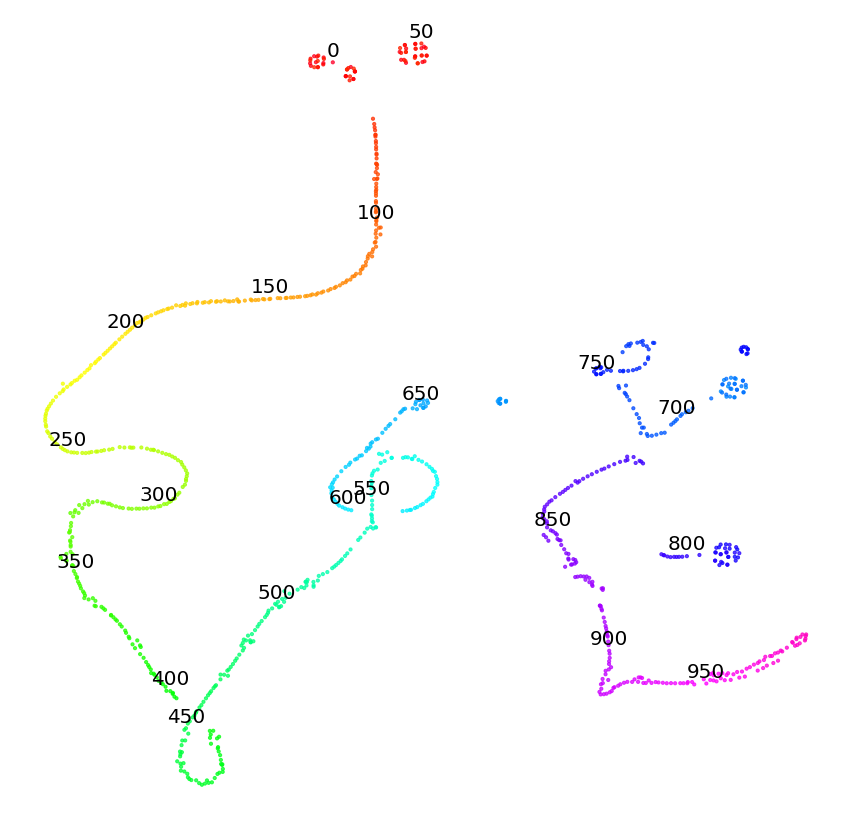}
\caption{Visualisation of the trajectory (1000 steps) of the agent in \texttt{MsPacman} environment. After training, the agent runs the obtained policy storing embeddings for each observation, produced by $E(\cdot)$. Then embedding are projected to 2D with the t-SNE algorithm. Nearby steps are depicted in a similar color, and the step counter shows specific points on the trajectory.}
\label{fig.path}
\end{figure}

Additionally to quantitative results, we include the visualisation of the resulting embeddings in Fig.~\ref{fig.path}. We use a trajectory of the agent after training in \texttt{MsPacman} environment. Each observation is encoded with $E(\cdot)$ and then projected to 2D with the t-SNE \cite{tsne} algorithm. We can see that observations are perfectly aligned and the encoder successfully learned the target representation. Interestingly, the curves of the path correspond to special events during the run. The game starts with an initial animation where the agent cannot move, so these frames (0 and 50) are bunched up. Similar animations are around points 700 and 800 after losing a life. After step 400 the agent takes a bonus which allows attacking opponents, this creates a gap with a twist in the path. Around step 550 the agent was trapped and it also results in a special twist.

It is important to note that our method requires an additional computational budget with respect to the baseline with only instantaneous input. The significant overhead comes from the temporal alignment stage and from producing embeddings for $H(t)$ during the policy optimisation stage. However, the encoder $E(\cdot)$ does not depend on the goal or the reward, so in a more realistic setting, we can train it once for the environment and reuse it for different tasks. In addition, if the encoder is not updated, the embeddings can be stored during experience collection, neglecting the corresponding overhead during the policy optimisation.

\subsection{Ablation study}
\label{ablation}

\begin{table}
\renewcommand{\arraystretch}{1.5}
\caption{Final averaged cumulative rewards of\\ablative versions of the agent}
\label{table.ablation}
\centering
\begin{tabular}{|l|c|c|c|c|}
\hline
Environment & Random & History	& Instant. & Ensemble \\
\hline
\hline
Asteroids & 810.5 & 1360.7 & 2157.7 & {\bf 2827.4}\\
Breakout & 1.3 & 30.48 & 370.14 & {\bf 409.42}\\
Freeway & 0 & 30.58 & {\bf 33.3} & 30.7\\
Frostbite & 74.7 & 276.1 & 293.8 & {\bf 848.2}\\
Gravitar & 194 & 448.5 & 455 & {\bf 642}\\
MsPacman & 237.6 & {\bf 1895.3} & 1840.9 & 1890.9\\
Seaquest & 80.8 & 778.4 & 1739.4 & {\bf 1843.6}\\
SpaceInvaders & 133.4 & 521.45 & 1070.9 & {\bf 1079}\\
Tennis & -23.9 & -2.31 & -9.43 & {\bf -1.49}\\
\hline
\end{tabular}
\end{table}

Most Atari environments are reactive, e.g. in \texttt{Breakout}, it is important to catch the ball precisely. In such cases, instantaneous input is favorable, while using only compact history may be not sufficient. We verify this assumption in the following experiment. We remove the instantaneous branch and corresponding input, and we use only the history branch. The only modification of hyperparameters is a bigger size of the FC layer $512$. Tab.~\ref{table.ablation} shows all the considered versions. The history-only method performs better than random, thus the agent was able to extract necessary information in every environment. In some cases the performance of history-only and instantaneous-only methods is similar, the history representation is sufficient. However, the ensemble shows the best result combining the advantages of both types of input.

\section{Conclusion}
We presented a method for history representation in RL. Our TempAl is based on the idea that important information about the past can be ``compressed'' using temporal alignment of observations. Specifically, this representation is learned using the common SSL approach based on the pairwise cross-entropy loss. The encoder and the agent are trained jointly and iterated through time, to adapt the history representation to the new observations.

In TempAl, visual information is represented using two different networks: the SSL encoder $E(\cdot)$ and the RL agent $\Theta$. The latter is trained using a reward signal, so it presumably extracts task-specific information from the observations. On the other hand, the encoder $E(\cdot)$ is trained using self-supervision, and thus it focuses on patterns that are repeated in the data stream, potentially leveraging a larger quantity of supervision signal. Although SSL has been explored in other RL and non-RL works, this is the first work to show how the discovered information can be exploited in the form of history representation.

{\bf Broader Impact.}
Our approach was demonstrated on Atari games, the most popular RL benchmark with image observations. This benchmark allows to run numerous simulations quickly, thus experiments can be carried out with a reasonable budget. However, simple sprite-based computer graphics, used in these games, is only a rough proxy to real-world problems. On the other hand, SSL has recently enjoyed remarkable success in the natural image domain \cite{simclr,byol}. Our representation learning is based on SSL, therefore TempAl can be used in environments with natural observations, e.g. in robotics. Furthermore, in some cases, TempAl can be pretrained from an unsupervised stream of observations, and later the encoder can be reused for other tasks in the environment since it does not depend on the reward. Moreover, real-world environments are often partially observable, e.g. in navigation, and the agent would benefit from the history input. We believe that our work can inspire further experimentation, advancing RL to practical applications.

\section*{Acknowledgment}
This work was supported by the EU H2020 AI4Media Project under Grant 951911.

\bibliographystyle{IEEEtran}
\bibliography{_main}

\begin{thebibliography}{10}
\providecommand{\url}[1]{#1}
\csname url@samestyle\endcsname
\providecommand{\newblock}{\relax}
\providecommand{\bibinfo}[2]{#2}
\providecommand{\BIBentrySTDinterwordspacing}{\spaceskip=0pt\relax}
\providecommand{\BIBentryALTinterwordstretchfactor}{4}
\providecommand{\BIBentryALTinterwordspacing}{\spaceskip=\fontdimen2\font plus
\BIBentryALTinterwordstretchfactor\fontdimen3\font minus
  \fontdimen4\font\relax}
\providecommand{\BIBforeignlanguage}[2]{{%
\expandafter\ifx\csname l@#1\endcsname\relax
\typeout{** WARNING: IEEEtran.bst: No hyphenation pattern has been}%
\typeout{** loaded for the language `#1'. Using the pattern for}%
\typeout{** the default language instead.}%
\else
\language=\csname l@#1\endcsname
\fi
#2}}
\providecommand{\BIBdecl}{\relax}
\BIBdecl

\bibitem{Mnih-2013}
V.~Mnih, K.~Kavukcuoglu, D.~Silver, A.~Graves, I.~Antonoglou, D.~Wierstra, and
  M.~Riedmiller, ``Playing atari with deep reinforcement learning,'' in
  \emph{NIPS Deep Learning Workshop}, 2013.

\bibitem{Mirhoseini2021}
A.~Mirhoseini, A.~Goldie, M.~Yazgan, J.~W. Jiang, E.~Songhori, S.~Wang, Y.-J.
  Lee, E.~Johnson, O.~Pathak, A.~Nazi, J.~Pak, A.~Tong, K.~Srinivasa, W.~Hang,
  E.~Tuncer, Q.~V. Le, J.~Laudon, R.~Ho, R.~Carpenter, and J.~Dean, ``A graph
  placement methodology for fast chip design,'' \emph{Nature}, vol. 594, no.
  7862, pp. 207--212, Jun 2021.

\bibitem{R2D2}
S.~Kapturowski, G.~Ostrovski, W.~Dabney, J.~Quan, and R.~Munos, ``Recurrent
  experience replay in distributed reinforcement learning,'' in \emph{ICLR},
  2019.

\bibitem{A3C}
V.~Mnih, A.~P. Badia, M.~Mirza, A.~Graves, T.~P. Lillicrap, T.~Harley,
  D.~Silver, and K.~Kavukcuoglu, ``Asynchronous methods for deep reinforcement
  learning,'' in \emph{ICML}, 2016.

\bibitem{DBLP:journals/corr/SchulmanMLJA15}
J.~Schulman, P.~Moritz, S.~Levine, M.~I. Jordan, and P.~Abbeel,
  ``High-dimensional continuous control using generalized advantage
  estimation,'' in \emph{ICLR}, 2016.

\bibitem{CPC}
A.~v.~d. Oord, Y.~Li, and O.~Vinyals, ``Representation learning with
  contrastive predictive coding,'' \emph{arXiv preprint arXiv:1807.03748},
  2018.

\bibitem{PCE}
M.~Boudiaf, J.~Rony, I.~M. Ziko, E.~Granger, M.~Pedersoli, P.~Piantanida, and
  I.~B. Ayed, ``A unifying mutual information view of metric learning:
  cross-entropy vs. pairwise losses,'' in \emph{European conference on computer
  vision}.\hskip 1em plus 0.5em minus 0.4em\relax Springer, 2020, pp. 548--564.

\bibitem{ST-DIM}
A.~Anand, E.~Racah, S.~Ozair, Y.~Bengio, M.-A. C{\^o}t{\'e}, and R.~D. Hjelm,
  ``Unsupervised state representation learning in atari,'' \emph{Advances in
  Neural Information Processing Systems}, vol.~32, 2019.

\bibitem{moco}
K.~He, H.~Fan, Y.~Wu, S.~Xie, and R.~Girshick, ``Momentum contrast for
  unsupervised visual representation learning,'' in \emph{Proceedings of the
  IEEE/CVF conference on computer vision and pattern recognition}, 2020, pp.
  9729--9738.

\bibitem{simclr}
T.~Chen, S.~Kornblith, M.~Norouzi, and G.~Hinton, ``A simple framework for
  contrastive learning of visual representations,'' in \emph{International
  conference on machine learning}.\hskip 1em plus 0.5em minus 0.4em\relax PMLR,
  2020, pp. 1597--1607.

\bibitem{PPO}
J.~Schulman, F.~Wolski, P.~Dhariwal, A.~Radford, and O.~Klimov, ``Proximal
  policy optimization algorithms,'' \emph{arXiv preprint arXiv:1707.06347},
  2017.

\bibitem{ALE}
M.~G. Bellemare, Y.~Naddaf, J.~Veness, and M.~Bowling, ``The arcade learning
  environment: An evaluation platform for general agents,'' \emph{Journal of
  Artificial Intelligence Research 47}, vol.~47, pp. 253--279, 2012.

\bibitem{Monahan}
G.~E. Monahan, ``State of the art{\textemdash}a survey of partially observable
  markov decision processes: Theory, models, and algorithms,'' \emph{Management
  Science}, vol.~28, no.~1, pp. 1--16, Jan. 1982.

\bibitem{Kaelbling}
L.~P. Kaelbling, M.~L. Littman, and A.~R. Cassandra, ``Planning and acting in
  partially observable stochastic domains,'' \emph{Artificial Intelligence},
  vol. 101, no.~1, pp. 99 -- 134, 1998.

\bibitem{DRQN}
M.~Hausknecht and P.~Stone, ``Deep recurrent q-learning for partially
  observable mdps,'' in \emph{2015 aaai fall symposium series}, 2015.

\bibitem{DBLP:journals/neco/HochreiterS97}
S.~Hochreiter and J.~Schmidhuber, ``Long short-term memory,'' \emph{Neural
  Computation}, vol.~9, no.~8, pp. 1735--1780, 1997.

\bibitem{DVRL}
M.~Igl, L.~Zintgraf, T.~A. Le, F.~Wood, and S.~Whiteson, ``Deep variational
  reinforcement learning for pomdps,'' in \emph{International Conference on
  Machine Learning}.\hskip 1em plus 0.5em minus 0.4em\relax PMLR, 2018, pp.
  2117--2126.

\bibitem{GRU}
K.~Cho, B.~van Merrienboer, C.~Gulcehre, D.~Bahdanau, F.~Bougares, H.~Schwenk,
  and Y.~Bengio, ``Learning phrase representations using rnn encoder-decoder
  for statistical machine translation,'' in \emph{EMNLP}, 2014.

\bibitem{byol}
J.-B. Grill, F.~Strub, F.~Altch{\'e}, C.~Tallec, P.~Richemond, E.~Buchatskaya,
  C.~Doersch, B.~Avila~Pires, Z.~Guo, M.~Gheshlaghi~Azar \emph{et~al.},
  ``Bootstrap your own latent-a new approach to self-supervised learning,''
  \emph{Advances in Neural Information Processing Systems}, vol.~33, pp.
  21\,271--21\,284, 2020.

\bibitem{W-MSE}
A.~Ermolov, A.~Siarohin, E.~Sangineto, and N.~Sebe, ``Whitening for
  self-supervised representation learning,'' in \emph{International Conference
  on Machine Learning}.\hskip 1em plus 0.5em minus 0.4em\relax PMLR, 2021, pp.
  3015--3024.

\bibitem{SimSiam}
X.~Chen and K.~He, ``Exploring simple siamese representation learning,'' in
  \emph{Proceedings of the IEEE/CVF Conference on Computer Vision and Pattern
  Recognition}, 2021, pp. 15\,750--15\,758.

\bibitem{barlow}
J.~Zbontar, L.~Jing, I.~Misra, Y.~LeCun, and S.~Deny, ``Barlow twins:
  Self-supervised learning via redundancy reduction,'' in \emph{International
  Conference on Machine Learning}.\hskip 1em plus 0.5em minus 0.4em\relax PMLR,
  2021, pp. 12\,310--12\,320.

\bibitem{NPair}
K.~Sohn, ``Improved deep metric learning with multi-class n-pair loss
  objective,'' in \emph{Advances in Neural Information Processing Systems},
  D.~Lee, M.~Sugiyama, U.~Luxburg, I.~Guyon, and R.~Garnett, Eds.,
  vol.~29.\hskip 1em plus 0.5em minus 0.4em\relax Curran Associates, Inc.,
  2016.

\bibitem{schwarzer2021dataefficient}
M.~Schwarzer, A.~Anand, R.~Goel, R.~D. Hjelm, A.~Courville, and P.~Bachman,
  ``Data-efficient reinforcement learning with self-predictive
  representations,'' \emph{arXiv preprint arXiv:2007.05929}, 2020.

\bibitem{world-models}
D.~Ha and J.~Schmidhuber, ``Recurrent world models facilitate policy
  evolution,'' in \emph{Advances in Neural Information Processing Systems},
  S.~Bengio, H.~Wallach, H.~Larochelle, K.~Grauman, N.~Cesa-Bianchi, and
  R.~Garnett, Eds., vol.~31.\hskip 1em plus 0.5em minus 0.4em\relax Curran
  Associates, Inc., 2018.

\bibitem{VAE}
D.~P. Kingma and M.~Welling, ``Auto-encoding variational bayes,'' in
  \emph{ICLR}, 2014.

\bibitem{LWM}
A.~Ermolov and N.~Sebe, ``Latent world models for intrinsically motivated
  exploration,'' in \emph{Advances in Neural Information Processing Systems},
  H.~Larochelle, M.~Ranzato, R.~Hadsell, M.~F. Balcan, and H.~Lin, Eds.,
  vol.~33.\hskip 1em plus 0.5em minus 0.4em\relax Curran Associates, Inc.,
  2020, pp. 5565--5575.

\bibitem{EffZero}
W.~Ye, S.~Liu, T.~Kurutach, P.~Abbeel, and Y.~Gao, ``Mastering atari games with
  limited data,'' \emph{Advances in Neural Information Processing Systems},
  vol.~34, 2021.

\bibitem{muzero}
J.~Schrittwieser, I.~Antonoglou, T.~Hubert, K.~Simonyan, L.~Sifre, S.~Schmitt,
  A.~Guez, E.~Lockhart, D.~Hassabis, T.~Graepel, T.~Lillicrap, and D.~Silver,
  ``Mastering atari, go, chess and shogi by planning with a learned model,''
  \emph{Nature}, vol. 588, no. 7839, pp. 604--609, Dec 2020.

\bibitem{sermanet2018timecontrastive}
P.~Sermanet, C.~Lynch, Y.~Chebotar, J.~Hsu, E.~Jang, S.~Schaal, S.~Levine, and
  G.~Brain, ``Time-contrastive networks: Self-supervised learning from video,''
  in \emph{2018 IEEE international conference on robotics and automation
  (ICRA)}.\hskip 1em plus 0.5em minus 0.4em\relax IEEE, 2018, pp. 1134--1141.

\bibitem{slow_feat}
L.~Wiskott and T.~J. Sejnowski, ``Slow feature analysis: Unsupervised learning
  of invariances,'' \emph{Neural Comput.}, vol.~14, no.~4, p. 715–770, 2002.

\bibitem{goroshin2015unsupervised}
R.~Goroshin, J.~Bruna, J.~Tompson, D.~Eigen, and Y.~LeCun, ``Unsupervised
  learning of spatiotemporally coherent metrics,'' 2015.

\bibitem{DBLP:conf/cvpr/HadsellCL06}
R.~Hadsell, S.~Chopra, and Y.~LeCun, ``Dimensionality reduction by learning an
  invariant mapping,'' in \emph{CVPR}, 2006.

\bibitem{ortho_init}
A.~M. Saxe, J.~L. McClelland, and S.~Ganguli, ``Exact solutions to the
  nonlinear dynamics of learning in deep linear neural networks,'' 2014.

\bibitem{tsne}
L.~van~der Maaten and G.~Hinton, ``Visualizing data using {t-SNE},''
  \emph{Journal of Machine Learning Research}, vol.~9, pp. 2579--2605, 2008.

\end{thebibliography}
\end{document}